\documentclass[10pt, a4paper]{article}

\usepackage{graphicx}
\usepackage{tabularx}
\usepackage{soul}
\usepackage{amsmath}
\usepackage{booktabs}
\usepackage{multirow}
\usepackage{subcaption}
\usepackage{tikz,lipsum,lmodern}
\usepackage[most]{tcolorbox}

\usepackage{lrec-coling2024} 

\title{Is it Possible to Modify Text to a Target Readability Level? An Initial Investigation Using Zero-Shot Large Language Models}

\name{Asma Farajidizaji\thanks{Asma Farajidizaji did the research in collaboration with the ALTA Institute}, Vatsal Raina, Mark Gales} 

\address{ALTA Institute, University of Cambridge, UK \\
         farajiasma@gmail.com, vr311@cam.ac.uk, mjfg@cam.ac.uk\\
         }

\abstract{
Text simplification is a common task where the text is adapted to make it easier to understand. Similarly, text elaboration can make a passage more sophisticated, offering a method to control the complexity of reading comprehension tests. However, text simplification and elaboration tasks are limited to only relatively alter the readability of texts. It is useful to directly modify the readability of any text to an absolute target readability level to cater to a diverse audience. Ideally, the readability of readability-controlled generated text should be independent of the source text. Therefore, we propose a novel readability-controlled text modification task. The task requires the generation of 8 versions at various target readability levels for each input text. We introduce novel readability-controlled text modification metrics.
The baselines for this task use ChatGPT and Llama-2, with an extension approach introducing a two-step process (generating paraphrases by passing through the language model twice). The zero-shot approaches are able to push the readability of the paraphrases in the desired direction but the final readability remains correlated with the original text's readability.
We also find greater drops in semantic and lexical similarity between the source and target texts with greater shifts in the readability.
 \\ \newline  }

\begin{document}

\maketitleabstract

\section{Introduction}

Natural language consists of information that is conveyed for a targeted audience. In order to make the text appropriate for a diverse set of readers, the source text needs to be modified accordingly. Automatic text simplification is a popular natural language processing (NLP) task where the source text is adapted to make the content easier to understand by reducing its linguistic complexity \citep{siddharthan2014survey, sikka2020survey}. Typically such simplification solutions are valuable for various audiences including younger readers \citep{de2010text}, foreign language speakers \citep{bingel2018lexi}, dyslexics \citep{rello2013frequent}, sufferers of autism \citep{evans2014evaluation} and aphasics \citep{carroll1998practical}. Similarly, text elaboration offers methods to make content more challenging for reading comprehension tasks and hence cater to higher level students \citep{ross1991simplification}.

However, both text simplification and elaboration are able to only relatively control the readability of the text. This means that the generated text is simplified/elaborated relative to the original text document but it does not guarantee the text itself is at an appropriate readability level for the target audience. In an ideal setting, it should be possible to modify a text document to a precise and absolute readability level. Pertinently, the readability of the modified text should be \textit{independent} and \textit{uncorrelated} with the source text's readability. Hence, regardless of the nature of the source text, it can be modified to any other readability level.

\begin{figure}[t]
    \centering
    \includegraphics[width=3.1in]{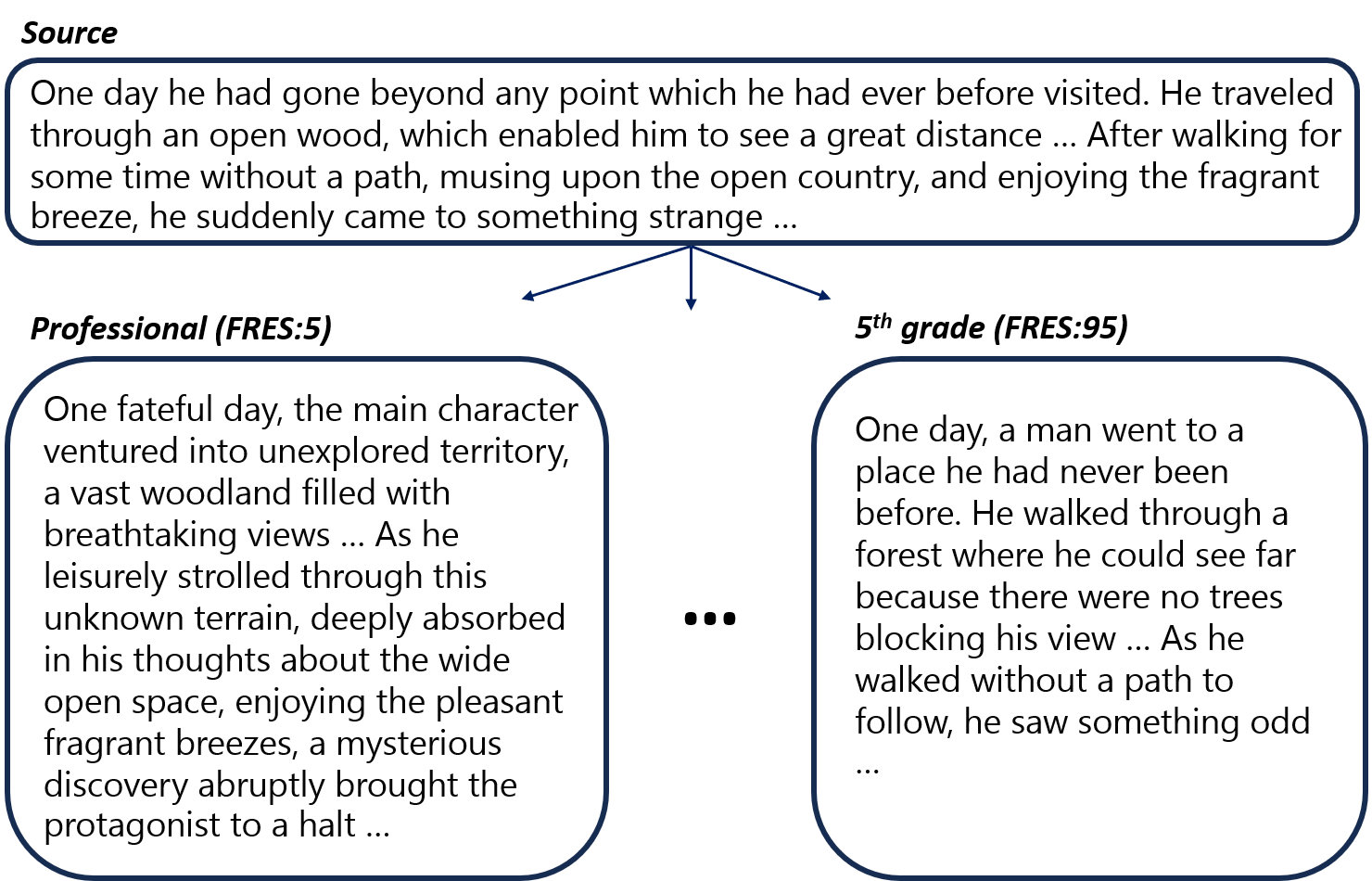}
    \caption{Example for the readability-controlled text modification task. The source text from CLEAR \citep{crossley2023large} is paraphrased at various target readability levels according to the Flesch reading ease score (FRES) \citep{flesch1948new}.}
    \label{fig:intro}
\end{figure}

To address the relative nature of current text modification approaches, we propose a novel text modification task to control text readability \citep{harris1995literacy}. Given a set of text documents across the whole spectrum of readability levels, generate 8 versions for each document corresponding to different target readability levels.
Precisely, the target readability scores are ranging from being readable for a $5^{\text{th}}$ grade student to understandable for university graduates \citep{flesch1948new}.

Paraphrasing is a common NLP task where a source text is modified to convey the same meaning but using different words or sentence structures \citep{zhou2021paraphrase}. Hence, automated paraphrasing solutions offer an opportunity to modify text to various target readability levels. However, standard solutions do not attempt to control the readability of the generated paraphrase and usually aim to maintain consistency with the source text \citep{kumar2020syntax, chen2020semantically}. Despite the lack of flexibility of these paraphrasing models, the remarkable growth of large-scale autoregressive foundation models \citep{zhou2023comprehensive} have demonstrated capabilities across a broad range of NLP tasks with simple prompting \citep{sanh2021multitask}. Thus, the baseline solutions for the novel task in our work use zero-shot \citep{brown2020language} prompting of such models as their backbones for readability-controlled paraphrasing. 

The text modification approaches that generate the eight adaptations of each source text document are assessed for their ability to control the readability. The proposed metrics assess both the readability control at an individual example level and the population level. At the individual scale, we assess, with various metrics, whether the readability of a text approaches the target value. At the population scale, we explore the extent to which the measured readability of a generated text document is conditional on the source text document's readability. Additionally, we explore the behaviour of the modified texts for each target readability level according to standard paraphrasing metrics. A good paraphrase can expect to be lexically divergent but semantically similar to the source.

Our contributions can be summarized as follows:
\begin{itemize}
    \item Introduction of a novel task for readability-controlled text modification.
    \item Definition of appropriate evaluation metrics for controlling readability.
    \item In-depth analysis of zero-shot large language model solutions for controlling text readability with paraphrasing.
\end{itemize}

\section{Related Work}

In this work we focus on controllability in text modification for readability. Previous works have explored similar approaches and tasks to control various attributes across a diverse range of natural language tasks. Here, we discuss the control of attributes in machine translation \citep{logeswaran2018content}, automatic summarization and text generation \citep{zhang2022survey}.

Machine translation is a natural language generation task that translates a source text into a different language. \citet{kikuchi2016controlling} investigates the ability to control the length of generated sentences such that translations can vary from brief summaries to longer texts. 
Beyond structural control, \citet{yamagishi2016controlling} controls the voice of the translation while \citet{sennrich2016controlling} controls the honorifics and politeness as the selected attributes.

Summarization is a standard natural language generation task where a source text must be condensed whilst maintaining the core elements of the original passage. Automatic summarization has observed the control of various attributes including length, entity-centric, source-specific and linked to a particular portion of the source document \citep{fan2018controllable}.

In text generation, \citet{zhang2022survey} states that the attributes to control are grouped into 3 distinct categories: semantic, structural and lexical.
Semantic control involves the control of emotion \citep{chen2019sentiment, dathathri2019plug} such as sentiment of the generated text as well the choice of the topic \citep{khalifa2020distributional} being discussed and the degree of toxicity \citep{krause2021gedi, liu2021dexperts} in the text.
Structural control typically looks at defining the syntax in the generated text and the occurrence of graphs and tables \citep{puduppully2019data, ribeiro2021investigating}. Finally, lexical control in text generation focuses on attributes such as the inclusion of keywords or phrases \citep{carlsson2022fine, he2021parallel}.

Besides controlling of attributes, there have been attempts to control text simplification to specific target readability levels. For example, \citet{alkaldi2023text} makes use of the Newsela dataset \citep{xu2015problems} to simplify challenging news articles to four different readability levels. In contrast, the work in this paper emphasises the need to be able to take text of any source readability to any target readability. Hence, text elaboration is important alongside text simplification. 


\section{Text Readability}

Text readability assesses how easy a piece of text is to read. Several standard measures exist for measuring the readability of text including the Flesch-Kincaid Grade Level \citep{kincaid1975derivation}, Dale Chall Readability \citep{dale1949concept}, Automated Readability Index (ARI) \citep{senter1967automated}, Coleman Liau Index \citep{coleman1975computer}, Gunning Fog \citep{gunning1952technique}, Spache \citep{spache1953new} and Linsear Write \citep{klare1974assessing}. 

In this work, the Flesch reading-ease \citep{flesch1948new} score (FRES) is used where higher scores indicate material that is easier to read while lower scores are reflective of more challenging passages. The score accounts for the ratio of the number of words to the number of sentences and the ratio of the number of syllables to the number of words to determine the overall readability as indicated in Equation \ref{eq:fres} \footnote{Implementation available at: \url{https://pypi.org/project/py-readability-metrics/}}.

\begin{equation}
\label{eq:fres}
\small
    \texttt{FRES} = 206.835 - 1.015\left( \frac{n_w}{n_{se}} \right) - 84.6\left( 
\frac{n_{sy}}{n_w} \right)
\end{equation}
where $n_w$ denotes the total number of words, $n_{se}$ denotes the total number of sentences and $n_{sy}$ denotes the total number of syllables.

\begin{table}[htbp!]
\centering
\small 
\begin{tabular}{ccp{3.5cm}}
\toprule
Range & Level (US) & Description \\
\midrule
0-10 & Professional & Extremely difficult to read. Best understood by university graduates. \\\\
10-30 & College graduate & Very difficult to read. Best understood by university graduates. \\\\
30-50 & College & 	Difficult to read. \\\\
50-60 & 10-12th grade & Fairly difficult to read. \\\\
60-70 & 8-9th grade & Plain English. Easily understood by 13- to 15-year-old students. \\\\
70-80 & 7th grade & Fairly easy to read. \\\\
80-90 & 6th grade & Easy to read. Conversational English for consumers. \\\\
90-100 & 5th grade & Very easy to read. Easily understood by an average 11-year-old student. \\
   \bottomrule
    \end{tabular}
\caption{Interpretable meaning of FRES \citep{flesch1948new}.}
\label{tab:fres}
\end{table}

FRES is selected as a simple measure for readability because it has highly interpretable ranges for the score as well as a high correlation with human comprehension as measured by reading tests \citep{dubay2007smart}. For example, Table \ref{tab:fres} shows that a FRES score below 10 indicates the text is readable by university graduates, FRES in the fifties is targeted for $10-12^{\text{th}}$ grade while FRES above 90 is readable for $5^{\text{th}}$ grade students. Such well defined ranges allows an exploration of the ability for controlling the readability of text. Note, FRES is not strictly constrained to be in the range of 0 to 100.

\section{Readability-Controlled Text Modification}

\subsection{Task definition}
\label{sec:task}

The readability-controlled text modification task is defined as follows: 
\\\\
``Given a text paragraph $x$, a function $\mathcal{F}$ for calculating readability scores, and $K$ pre-defined readability scores $r_1$, $r_2$, ..., $r_K$, generate $K$ versions of $x$ ($y_1$, $y_2$, ..., $y_K$), such that $\mathcal{F}(y_1) = r_1$, $\mathcal{F}(y_2) = r_2$, ..., $\mathcal{F}(y_K) = r_K$.''
\\\\
In this work, $K=8$ with $r_1=5$, $r_2=20$, $r_3=40$, $r_4=55$, $r_5=65$, $r_6=75$, $r_7=85$, $r_8=95$ and FRES (see Equation \ref{eq:fres}) is selected as the readability function, $\mathcal{F}$. This task is applied for every text in a dataset of text paragraphs. The target readability scores are selected as the halfway values for each range of FRES from Table \ref{tab:fres}.

\subsection{Evaluation}
\label{sec:evaluation}

The quality of the readability-controlled text modifications generated are assessed according to individual and population scale control in readability as well as additional analysis with standard paraphrasing metrics. 
\newline\newline
\textbf{Individual-scale readability control}:
For each example in a test set, 8 paraphrases are generated. The individual-scale readability control metrics assess the ability to appropriately control the readability of these paraphrases for each individual example. Broadly, the ranking, regression and classification abilities of a readability-controlled paraphrase generator are assessed.

Let $x$ denote the original text sequence, $y_{(r)}$ denote the generated paraphrase with target readability score of $r\in \mathcal{R} = \{5,20,40,55,65,75,85,95\}$. Let $\mathcal{F}(\cdot)$ represent the function for calculating FRES from Equation \ref{eq:fres}.

The ranking ability is assessed by calculating the Spearman's rank correlation coefficient, $\rho$, between the 8 values of $\mathcal{F}\left(y_{(r\in\mathcal{R})}\right)$ and $\mathcal{R}$. Hence, here we only assess whether the order of the generated paraphrases aligns with their target readabilities.

Given the target readability scores, the regression ability of the model is assessed by calculating the root mean square error (rmse) between the actual and target readability scores of the paraphrases.
\begin{equation}
    \text{rmse} = \left[\frac{1}{8}\sum_{r\in\mathcal{R}} (\mathcal{F}(y_{(r)}) - r)^2 \right]^{1/2}
\end{equation}

Finally, the classification ability checks the ability of the paraphrase generator to control the readability of the generated text into the target range as defined in Table \ref{tab:fres}. For example, a paraphrase with a target readability of 65 is deemed correct if the measured generated text readability is in the range of 60-70 and incorrect otherwise. Therefore, the classification accuracy can be calculated according to Equation \ref{eq:acc}.
\begin{equation}
\label{eq:acc}
    \text{accuracy} = \frac{1}{8}\sum_{r\in\mathcal{R}}\mathbf{1}_{\mathcal{A}_{(r)}}(\mathcal{F}(y_{(r)}))
\end{equation}
where $\mathcal{A}_{(5)} \in [0,10]$, $\mathcal{A}_{(20)} \in [10,30]$, $\mathcal{A}_{(40)} \in [30,50]$, $\mathcal{A}_{(55)} \in [50,60]$, $\mathcal{A}_{(65)} \in [60,70]$, $\mathcal{A}_{(75)} \in [70,80]$, $\mathcal{A}_{(85)} \in [80,90]$, $\mathcal{A}_{(95)} \in [90,100]$.

For the ranking, regression and classification metrics, the mean is reported across the test set of examples.
\newline\newline
\textbf{Population-scale readability control}
These metrics assess the actual readability of each target readability across a whole population (test set) rather than considering each example individually. In particular, an important aspect of readability control requires the controlled readability of the generated text to be decorrelated and independent with the source passage readability. In principle, the original text should not have any influence on the readability of the generated text if the control of the paraphrase generator is ideal.

First, we report the Pearson's correlation coefficient (pcc) between the source readability and the calculated generated text readability separately for each target readability class. Ideally, a decorrelated score should expect pcc=0.

Additionally, a linear regression line of the form of $y=ax+b$ is calculated for each target readability class between the source and generated text readability scores. In an ideal setting, the regression line should approach a gradient $a=0$. 
\newline\newline
\textbf{Standard paraphrasing}
A good paraphrase should be lexically divergent but semantically similar to the original text \citep{gleitman1970phrase, chen2011collecting, bhagat2013paraphrase}. In line with \citet{lin2021towards}, we assess lexical divergence using
self-WER \citep{och2003minimum} \footnote{Implementation available at: \url{https://github.com/belambert/asr-evaluation}}. Semantic similarity is assessed using BERTScore \citep{zhang2019bertscore} \footnote{Implementation available at: \url{https://github.com/Tiiiger/bert_score}}.

Self-WER calculates the word error rate (WER) (inspired from automatic speech recognition \citep{malik2021automatic} and machine translation \citep{lee2023survey}) between the source and generated text respectively. A lexically divergent paraphrase can expect to have a high self-WER. BERTScore compares the semantic similarity of the source and paraphrase by calculating the pairwise cosine similarities between pre-computed BERT \citep{kenton2019bert} token embeddings of each of the texts. Hence, the F1 metric is reported as the harmonic mean of precision and recall. 


\section{Experiments}

\subsection{Data}

CLEAR \citep{crossley2023large, crossley2021commonlit} is a large-scaled corpus for assessing text readability. Here, it is used as a test set of input passages on which readability-controlled text modification is performed. Table \ref{tab:clear} outlines the main statistics. There are roughly 5000 different texts with a mean of 10 sentences, allowing the text modification task to be performed at the passage-level rather than at the sentence-level.

\begin{table}[htbp!]
\centering
\small 
\begin{tabular}{cccc}
\toprule
\# examples & \# words & \# sentences & \# paragraphs \\
\midrule
4,724 & $179_{\pm 18}$ & $9.6_{\pm 4.6}$ & $2.5_{\pm 1.9}$ \\
   \bottomrule
    \end{tabular}
\caption{CLEAR dataset statistics.}
\label{tab:clear}
\end{table}

\begin{figure}[t]
    \centering
    \includegraphics[width=2.5in]{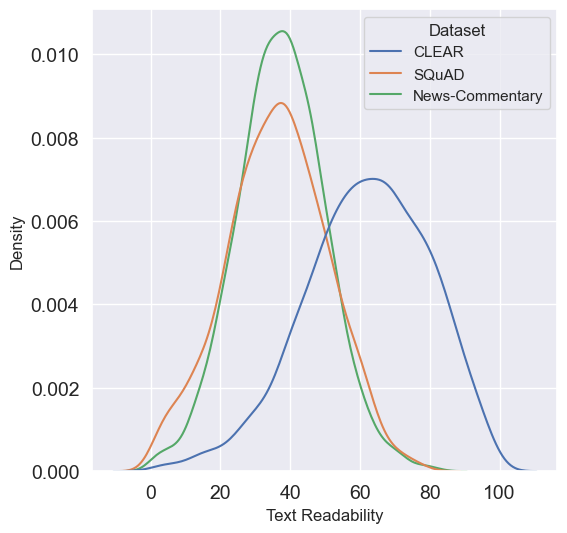}
    \caption{Distribution of text readability scores.}
    \label{fig:data}
\end{figure}

Other standard datasets exist for text simplification but these are generally at the sentence-level \citep{sun2021document} while we focus on longer texts. Alternatively, various popular passage-level datasets exist in reading comprehension and paraphrasing literature.
Figure \ref{fig:data} compares the distribution of the FRES for the passages within CLEAR, the SQuAD \citep{rajpurkar2016squad} development set and News-Commentary \citep{lin2021towards} test set. Due to the presence of texts across the whole spectrum of FRES scores, CLEAR is an attractive choice for investigating readability-controlled text modification. Hence, the experiments here are conducted on the CLEAR dataset only.

\subsection{Zero-shot}

\label{sec:models}

\begin{table}[htbp!]
\centering
\small 
\begin{tabular}{cp{6.0cm}}
\toprule
Target & Prompt \\
\midrule
5 & Paraphrase this document for a professional. It should be extremely difficult to read and best understood by university graduates. \\\\
20 & Paraphrase this document for college graduate level (US). It should be very difficult to read and best understood by university graduates. \\\\
40 & Paraphrase this document for college level (US). It should be difficult to read. \\\\
55 & Paraphrase this document for 10th-12th grade school level (US). It should be fairly difficult to read. \\\\
65 & Paraphrase this document for 8th/9th grade school level (US). It should be plain English and easily understood by 13- to 15-year-old students. \\\\
75 & Paraphrase this document for 7th grade school level (US). It should be fairly easy to read. \\\\
85 & Paraphrase this document for 6th grade school level (US). It should be easy to read and conversational English for consumers. \\\\
95 & Paraphrase this document for 5th grade school level (US). It should be very easy to read and easily understood by an average 11-year old student. \\
   \bottomrule
    \end{tabular}
\caption{Model prompts for each target readability level.}
\label{tab:prompts}
\end{table}

\begin{figure*}[t]
    \centering
    \begin{subfigure}[t]{0.66\columnwidth}
        \centering
        \includegraphics[width=2.0in]{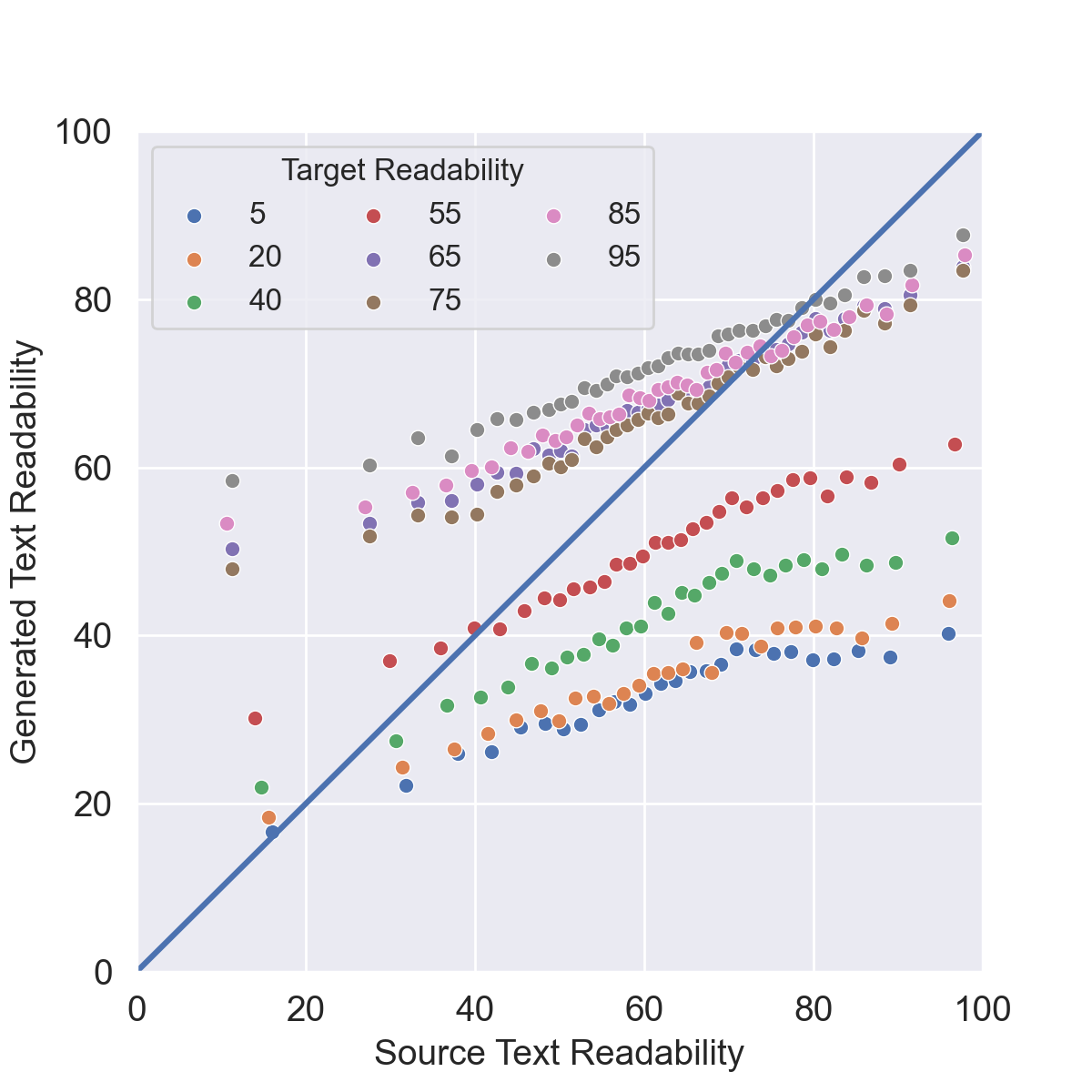}
        \caption{ChatGPT 1-step}
    \end{subfigure}%
    ~ 
    \begin{subfigure}[t]{0.66\columnwidth}
        \centering
        \includegraphics[width=2.0in]{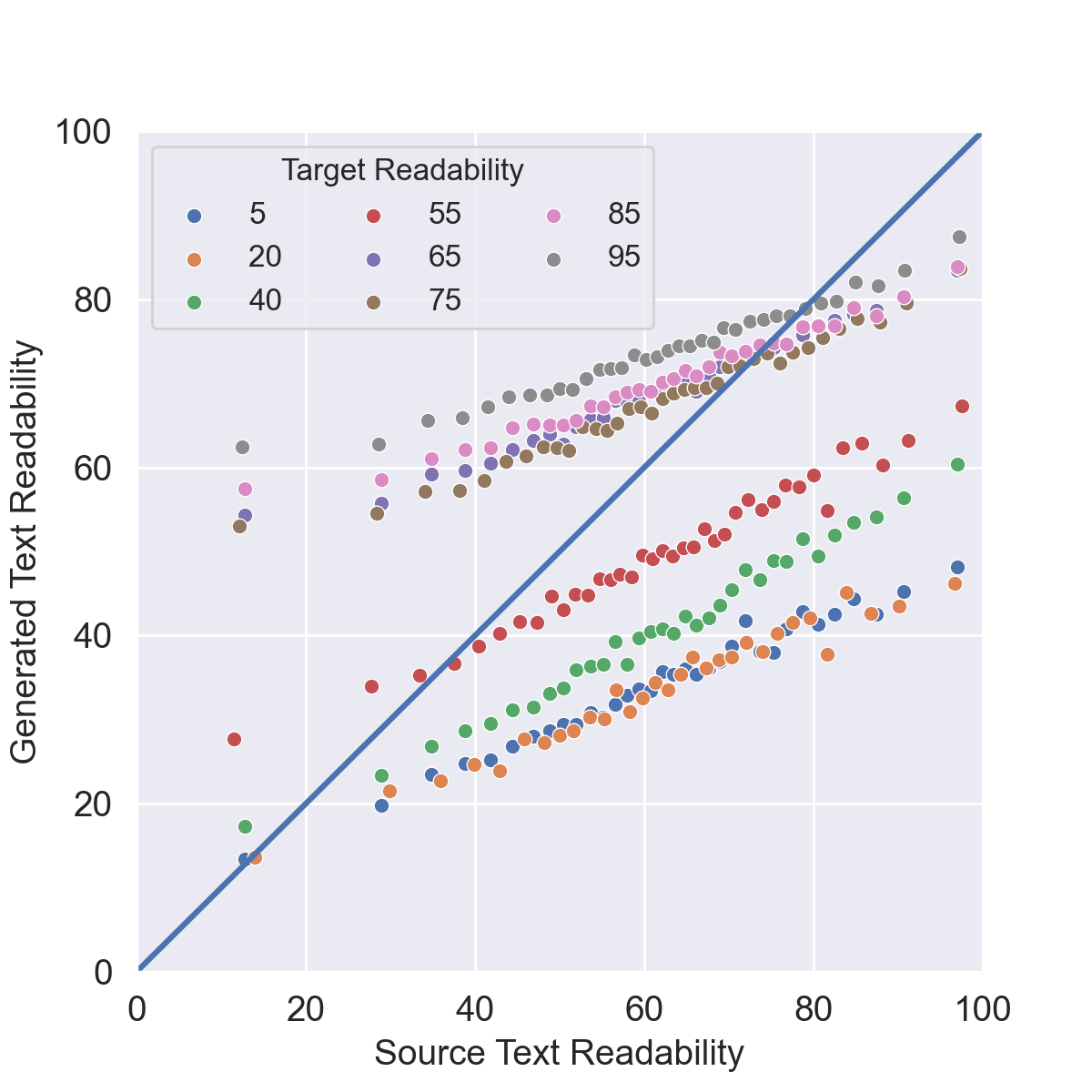}
        \caption{ChatGPT 2-step}
    \end{subfigure}%
    ~
    \begin{subfigure}[t]{0.66\columnwidth}
        \centering
        \includegraphics[width=2.0in]{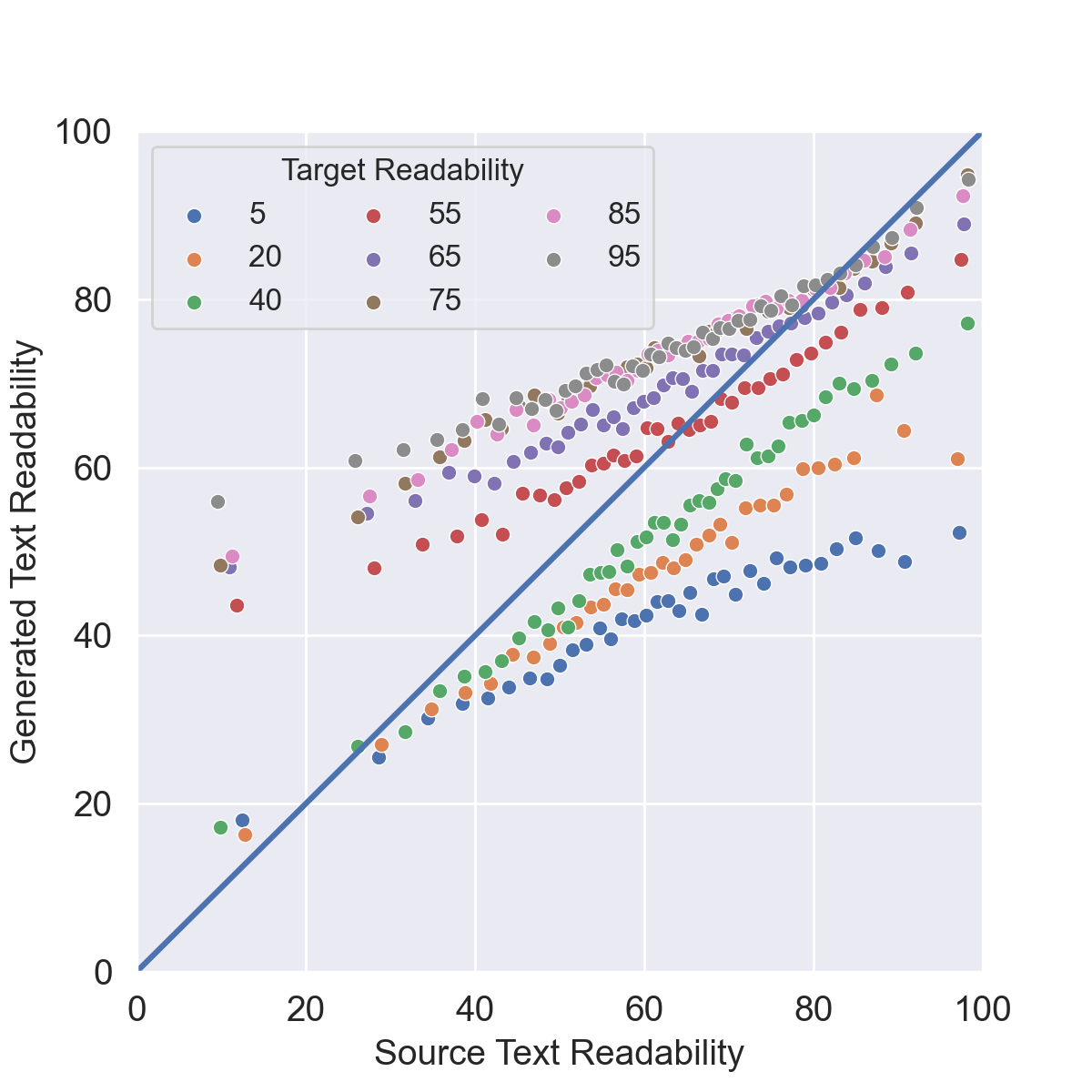}
        \caption{Llama-2}
    \end{subfigure}%
    \caption{Generated text readability against source text readability as a binned scatterplot.}
    \label{fig:binned_hist}
\end{figure*}

Large-scale generative foundation models \citep{brown2020language, chowdhery2022palm, scao2022bloom}, including the popularized ChatGPT, have demonstrated state-of-the-art performance across a large range of natural language tasks in zero-shot and few-shot settings. Despite not having been specifically trained on certain tasks, these models are capable of successfully performing novel tasks with natural language prompting. Therefore, our baseline solutions for readability-controlled text modification involve zero-shot solutions using ChatGPT and Llama-2 \citep{touvron2023llama}. Specifically, we use \texttt{gpt-3.5-turbo} \footnote{API access through \url{https://platform.openai.com/docs/models/gpt-3-5}} and \texttt{Llama-2-7b-chat-hf} \footnote{Available at: \url{https://huggingface.co/meta-llama/Llama-2-7b-chat-hf}} respectively.

\begin{table*}[t]
\centering
\small 
\begin{tabular}{l|ccc}
\toprule
Approach  & $\rho$ $(\uparrow)$ & rmse $(\downarrow)$ & accuracy $(\uparrow)$ \\
\midrule
Copy & $0.0$ & $35.4$ & $12.5$ \\
ChatGPT 1-step & $\mathbf{87.5}_{\pm 9.2}$ & $19.4_{\pm 4.9}$ & $23.1_{\pm 12.9}$ \\
ChatGPT 2-step & $86.0_{\pm 9.0}$ & $\mathbf{19.2}_{\pm 4.5}$ & $\mathbf{24.2}_{\pm 13.2}$ \\
Llama-2 & $73.3_{\pm 24.1}$ & $23.6_{\pm 8.0}$ & $20.6_{\pm 12.6}$ \\
   \bottomrule
    \end{tabular}
\caption{Individual-scale readability control with Spearman's rank correlation coefficient (as \%), $\rho$, regression ability with rmse and classification accuracy. The mean across all the examples is reported for each performance metric as well as one standard deviation.}
\label{tab:control_ind}
\end{table*}

\begin{table*}[t]
\centering
\small 
\begin{tabular}{c|cccc|cccc|cccc}
\toprule
& \multicolumn{4}{c|}{ChatGPT 1-step} & \multicolumn{4}{c|}{ChatGPT 2-step} & \multicolumn{4}{c}{Llama-2} \\
Target & pcc $(\downarrow)$ & $a$ $(\downarrow)$ & $b$ & $r^2$ & pcc $(\downarrow)$ & $a$ $(\downarrow)$ & $b$ & $r^2$ & pcc $(\downarrow)$ & $a$ $(\downarrow)$ & $b$ & $r^2$ \\
\midrule
Source & 100 & 1 & 0 & 1 & 100 & 1 & 0 & 1 & 100 & 1 & 0 & 1 \\
5 & 36.9 & 0.29 & 14.8 & 0.14 & 52.3 & 0.41 & 8.7 & 0.27 & 47.7 & 0.41 & 16.4 & 0.23 \\
20 & 39.9 & 0.32 & 15.3 & 0.16 & 50.8 & 0.40 & 8.8 & 0.26 & 51.1 & 0.61 & 9.9 & 0.26 \\
40 & 46.0 & 0.38 & 18.2 & 0.21 & 58.6 & 0.52 & 8.2 & 0.34 & 70.0 & 0.73 & 6.9 & 0.49 \\
55 & 56.0 & 0.41 & 24.7 & 0.31 & 62.2 & 0.47 & 20.4 & 0.39 & 65.1 & 0.51 & 33.0 & 0.42 \\
65 & 69.7 & 0.42 & 41.7 & 0.49 & 65.8 & 0.38 & 45.9 & 0.43 & 67.2 & 0.48 & 32.9 & 0.45 \\
75 & 68.8 & 0.44 & 39.4 & 0.47 & 65.0 & 0.38 & 44.2 & 0.42 & 62.6 & 0.47 & 44.3 & 0.39 \\
85 & 67.8 & 0.39 & 44.9 & 0.46 & 63.4 & 0.33 & 49.6 & 0.40 & 63.5 & 0.47 & 44.4 & 0.40 \\
95 & 66.3 & 0.36 & 50.3 & 0.44 & 61.3 & 0.31 & 54.8 & 0.38 & 61.3 & 0.42 & 48.1 & 0.38 \\
   \bottomrule
    \end{tabular}
\caption{Population-scale readability control with Pearson's correlation coefficient, pcc,  and linear regression, $y=ax+b$, between the source and generated text readability with $0<r^2<1$ denoting the quality of the fit of the regression line with 0 as worst fit and 1 as best fit.}
\label{tab:control_pop}
\end{table*}

Note, Llama-2 model weights are open-sourced, allowing future solutions to further finetune the zero-shot solution specifically for readability controlled text modification. Inference with ChatGPT only requires API requests while for Llama-2, generating 8 paraphrases per example passage takes approximately 45 seconds on an Nvidia A100 GPU. All experiments conducted in this work are based on publicly accessible datasets and models for reproducibility \footnote{Experiments available at: \url{https://github.com/asma-faraji/text-readability-control}}.
\newline\newline
\textbf{Vanilla}
The zero-shot solutions using ChatGPT and Llama-2 require natural language prompts to control the generated paraphrases. As the models do not have an inherent understanding of the FRES, explicit prompts are required to control the readability appropriately. Table \ref{tab:prompts} summarizes the prompts corresponding to each target readability level as defined in Section \ref{sec:task}. The prompts are selected in relation to the descriptions in Table \ref{tab:fres}.

It is observed that often the outputs from the Llama-2 zero-shot solution for certain input passages are random incoherent string of tokens. Therefore, a simple garbage detector checks whether the generated paraphrase is coherent English and in the situation garbage is detected, the corresponding paraphrase is replaced by the original text \footnote{This has only been observed to occur in under 1\% of generated paraphrases.}.
\newline\newline\noindent 
\textbf{Two-step}
Unlike many other natural language generation tasks (such as question generation \citep{lu2021survey}, summarization \citep{widyassari2022review} and question-answering \citep{baradaran2022survey}), the nature of the output matches the input for text modification. Therefore, paraphrasing based zero-shot approaches to control readability using large language models can sequentially be applied multiple times on a source text. Here, the two-step process is as follows: 1. the selected large language model is prompted to generate a paraphrase at the target readability level as according to Table \ref{tab:prompts} with the source text at the input; 2. the model is then again prompted (with the identical prompt) to generate a new text but instead with the output from the previous step at the input. The intuition for this approach is motivated by the concept that it is possible to shift closer to a target readability if the source readability is closer to the target value. Here, we explore the two-step process for ChatGPT as it's the higher performing model (see Table \ref{tab:control_ind}).

\section{Results and Discussion}

As described in Section \ref{sec:models}, several baseline solutions are considered for the readability controlled to target values. Table \ref{tab:control_ind} presents the performance of these solutions for the individual-scale metrics averaged across all examples in the CLEAR test set (see Section \ref{sec:evaluation}). The \textit{copy} system represents the setup where the source text is simply copied for each of target readability levels 5, 20, 40, 55, 65, 75, 85 and 95. Hence, the \textit{copy} system offers a lowerbound on performance according to each of the metrics.

According to the Spearman's rank correlation coefficient, all ChatGPT and Llama-2 implementations are effective at relatively controlling the readability of the text documents with ChatGPT 1-step attaining the highest correlation of 87.5\% while Llama-2 lags behind by about 15\%. In contrast, the models struggle to directly map the readability of texts to absolute target readability levels with rmse values spanning typically two readability ranges (see Table \ref{tab:fres}) and the classification accuracies below 25\%. Given all approaches are zero-shot implementations where the language models do not have an exact understanding of FRES, it is understandable the models are able to achieve a sensible readability ranking for the 8 generated texts but are incapable of matching the exact target readability values. Additionally, it can be noted that the 2-step process on ChatGPT observes incremental improvements (roughly 0.2 rmse and 1.0 classification accuracy) in achieving the absolute target readability values compared to the 1-step process. This perhaps is because there are two attempts to push the model towards the desired numeric readability score.

\begin{figure*}[t]
    \centering
    \begin{subfigure}[t]{0.9\columnwidth}
        \centering
        \includegraphics[width=2.5in]{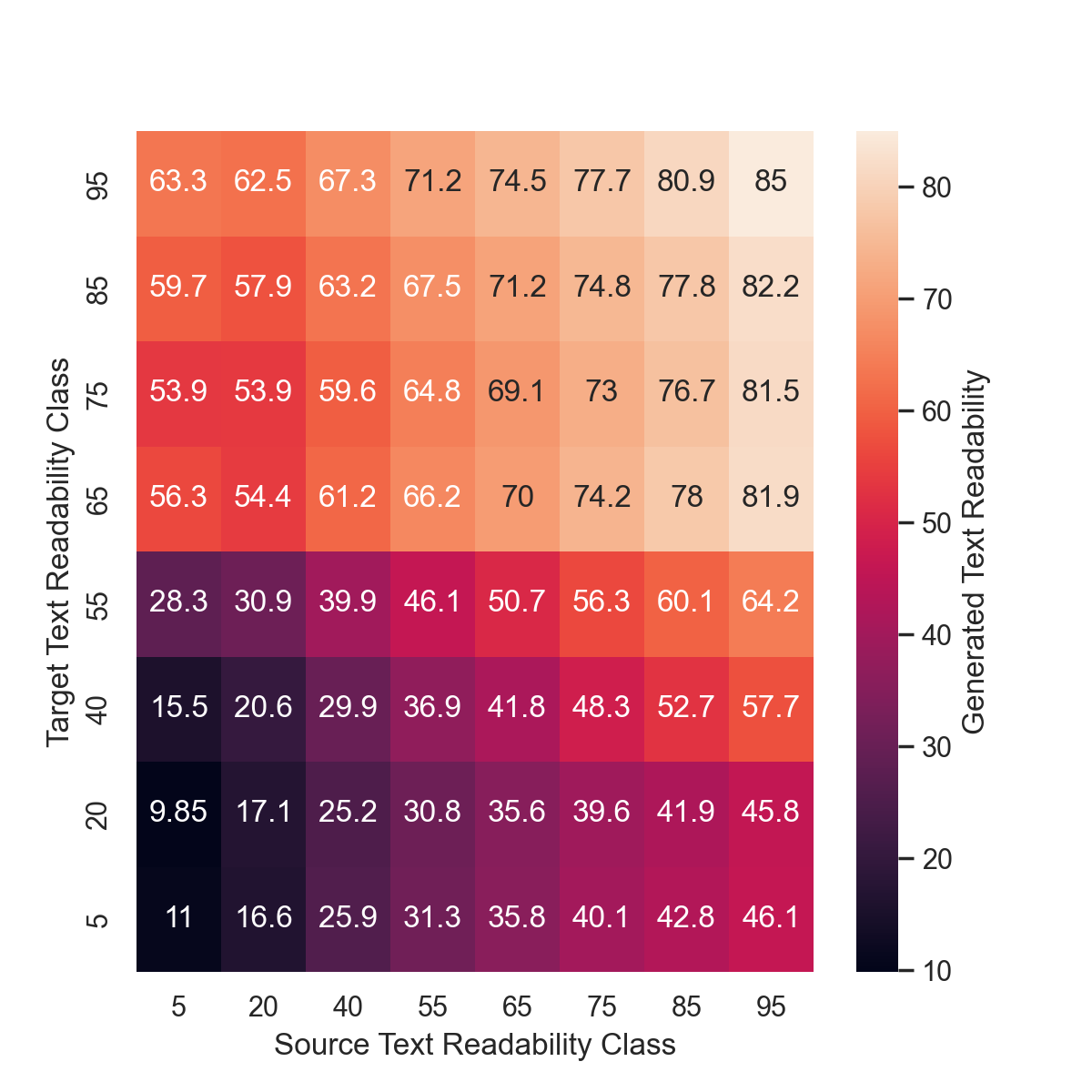}
        \caption{Generated FRES readability}
        \label{fig:heat_gen}
    \end{subfigure}%
    ~ 
    \begin{subfigure}[t]{0.9\columnwidth}
        \centering
        \includegraphics[width=2.5in]{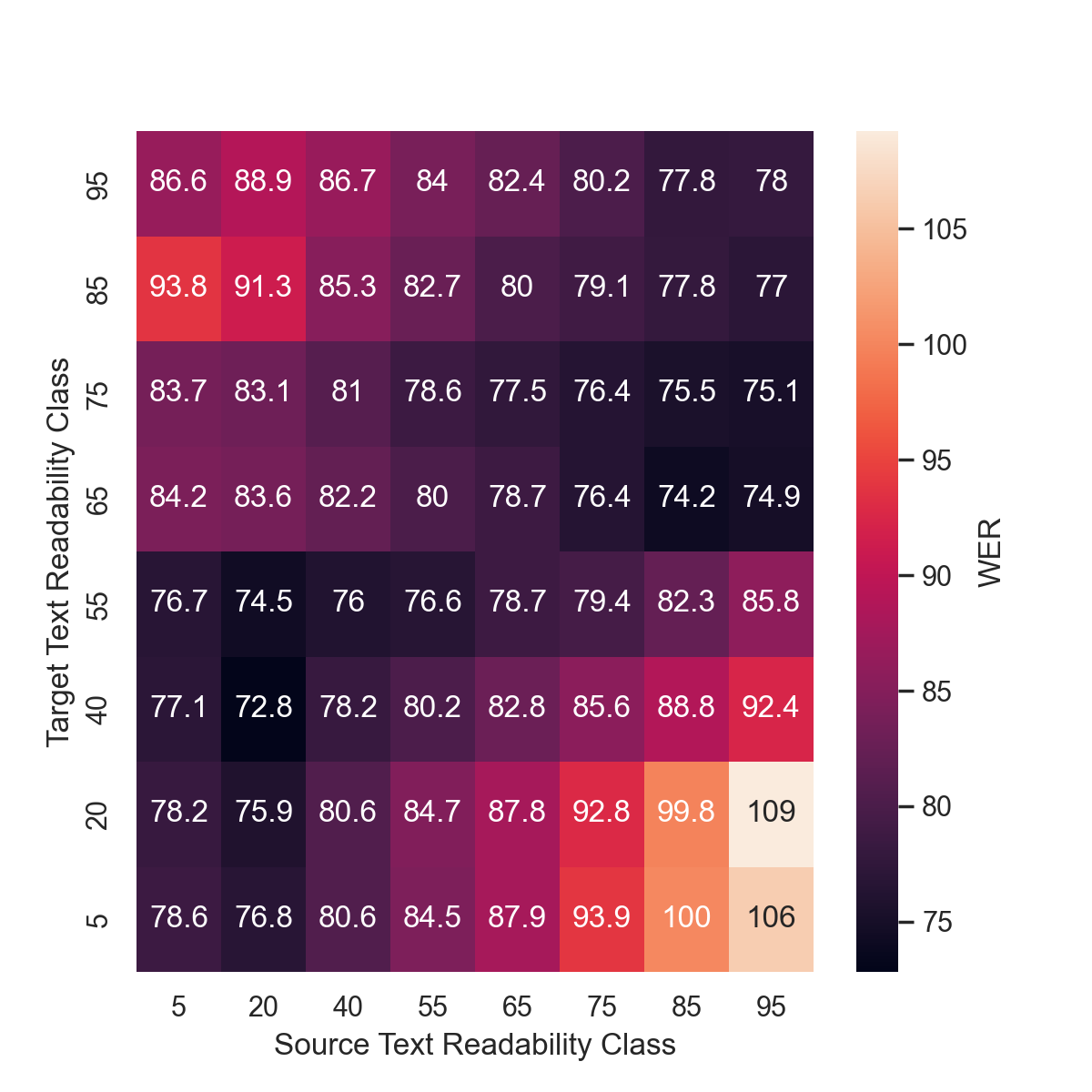}
        \caption{WER}
        \label{fig:heat_wer}
    \end{subfigure}%
    \\
    ~
    \begin{subfigure}[t]{0.9\columnwidth}
        \centering
        \includegraphics[width=2.5in]{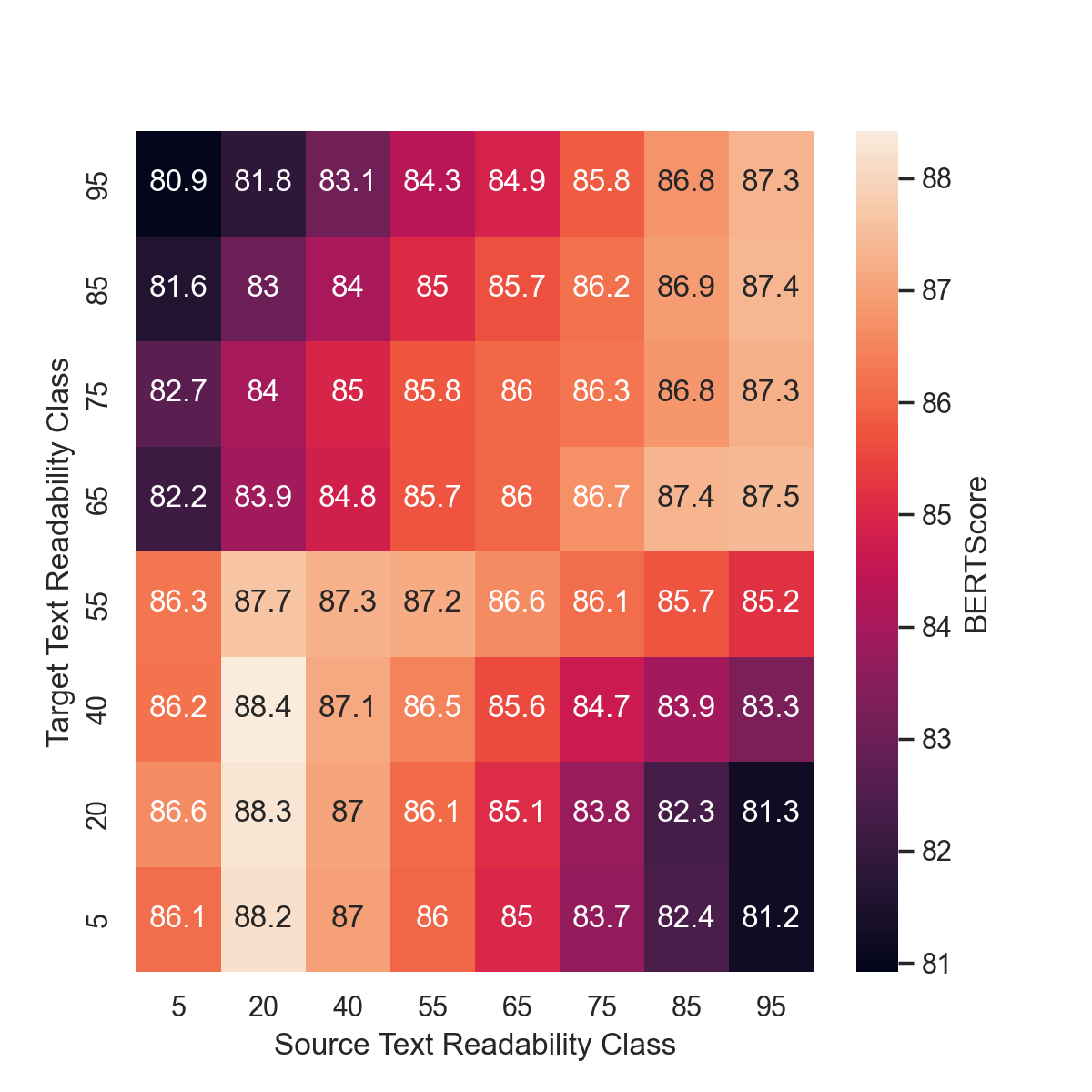}
        \caption{BERTScore F1}
        \label{fig:heat_bert}
    \end{subfigure}%
    ~
    \begin{subfigure}[t]{0.9\columnwidth}
        \centering
        \includegraphics[width=2.3in]{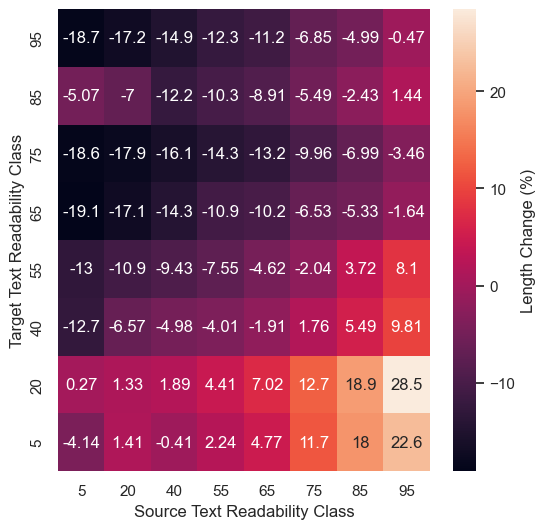}
        \caption{Length change (words)}
        \label{fig:heat_len}
    \end{subfigure}%
    \caption{Heatmaps of averaged select variables for each pair of source and target text readability classes. Each cell value is the mean of the specified variable for all texts that have a certain source readability and are modified to a certain target readability. }
    \label{fig:heatmap}
\end{figure*}

Figure \ref{fig:binned_hist} presents the relationship between the source text readability and the generated text measured readability for each of the target readability classes 5 to 95. The relationship is plotted as a binned scatterplot where the average measured readability is plotted for each bin on the source text readabilities. For all 3 models, it is observed that the readabilities of each class are generally in a sensible order where the measured readabilities for the target class 5 run along the bottom and the scores for the target class 95 act as the highest curve. Llama-2 also appears to be better than ChatGPT at disentangling the 5 and 20 classes but struggles at the higher classes. Additionally, the 2-step process is able at the lower target readabilities to push down to lower measured readabilities compared to the 1-step process.

However, it is also apparent from Figure \ref{fig:binned_hist} that the measured readability of the generated texts, albeit correctly ordered, is highly correlated with the source text readability. Table \ref{tab:control_pop} further quantifies the ability of the models to decorrelate the measured readability of the generated text with the source text readability according to the population-scale metrics (see Section \ref{sec:evaluation}). An ideal system can expect to have a Pearson's correlation coefficient of 0 and a regression line of best fit with gradient 0 and y-intercept corresponding to the absolute target readability level. It is observed that the relationship for all target readability classes remains highly dependent with the source text readability with ChatGPT 1-step achieving best results for lower target readability classes and better results for ChatGPT 2-step and Llama-2 for higher readability classes. It is also seen that the lower target readability classes are closer to the ideal performance for all models compared to the higher target readability classes.


For further analysis, we look at the behaviour of the generated texts at various target readability metrics according to lexical divergence and semantic similarity from paraphrasing literature (see Section \ref{sec:evaluation}). We present the analysis here specifically for the best performing model overall: ChatGPT 2-step.

Figures \ref{fig:heat_gen}, \ref{fig:heat_wer} and \ref{fig:heat_bert} display how the generated text readability, WER (measure of lexical divergence) and BERTScore F1 (measure of semantic similarity) of the generated texts respectively vary with shifts between the target and source readability score classes.
In order to plot the heatmap, each source text has its readability classed into one of the 8 readability ranges defined by Table \ref{tab:fres}. Hence the source and target text readability classes fall into one of following classes: $\{5,20,40,55,65,75,85,95\}$.
The heatmap depicts the mean of the selected variable (the variable for the heatmap to plot includes either generated text readability, WER or BERTScore) for each pairing of going from the source readability class (several source texts fall into each class) and the corresponding target readability class.

First, we see Figure \ref{fig:heat_gen} reinforces the observations from Figure \ref{fig:binned_hist} as the lightest colours are observed in the top right while the darkest in the bottom left. This means that the highest generated text readability scores are observed for when the target readability class is high but also when the source text readability is high.

From Figure \ref{fig:heat_wer} it is noticeable that along the leading diagonal we have the darkest shades and lighter shades on the peripheries of the off-diagonal. Conversely, from Figure \ref{fig:heat_bert} we see the lighter shades in the leading diagonal. Hence, keeping a matched readability between the source and the target leads to lower lexical divergence and higher semantic similarity.
It can further be noted that there is an asymmetry for the WER. For the WER, we see that changing from a very high source text readability to a very low target readability has a greater WER compared to a modification from a low source text readability to a high text readability. This suggests that it is more challenging to maintain the same lexical language for text elaboration compared to text simplification. 

We can conclude from the variations in WER and BERTScore that greater the change between the source and target texts, the lower their semantic similarity and greater their lexical divergence.

Additionally, Figure \ref{fig:heat_len} presents the percentage change in the length of the paraphrase relative to the original text for each binned pair of source and target readability levels. The resulting heatmap is in agreement with the observations from Figure \ref{fig:heat_wer} where the least absolute change in length is on the leading diagonal while greater variations in lengths are observed when the paraphrasing model is requested to make a greater change in the readability. Moreover, Figure \ref{fig:heat_len} presents a further asymmetry in the sign of the length change. It is observed that length changes are positive in the bottom right corner while the length change is negative in the top left. Hence, decreasing the readability of the text leads to an increase in the overall text length while requesting an increase in the readability causes a reduction in the text length.

\section{Ablations}

The core results focus on zero-shot approaches. In this section, an ablation study investigates the potential of finetuning a system for readability controlled text modification. Note, finetuning a system on the task enables absolute understanding of the readability measure, which is not inherently possible with zero-shot approaches.

Here, we look at finetuning Llama-2. Hence, we partition the CLEAR dataset into train and test splits with 2,834 and 1,890 examples respectively. For each text passage in the train split, we generate 8 paraphrases at the set target readability levels described in the core results with ChatGPT. This allows us to form a labeled training dataset of input-output pairs where we have annotations for the source readability for the source passage and the measured readability (not the target readability) for each generated passage. Then Llama-2 is finetuned according to the following prompt:
\newline\noindent
\texttt{\{source\_text\}
Paraphrase the following document changing the readability from
\{source\_score\} to \{target\_score\}}.

We use parameter efficient finetuning (due to compute limits) with quantized low rank adapters (QLoRA) \citep{dettmers2023qlora}. Final hyperparamters include: learning rate of 1e-3, batch size of 4, lora rank of 8, lora $\alpha$ of 32 and dropout 0.1. Training takes 2 hours for 3 epochs on NVIDIA A100.

The finetuned system achieves a Spearman's rank correlation of 61.4 (scaled by 100), rmse of 29.4 and accuracy of 16.3\%. Despite the Llama-2 finetuned system being the worst performing, it has the most exciting results because it is the only model that actually understands the quantitative meaning of the readability score. Hence, it offers opportunity for improvement while the zero-shot models can only be prompted qualitatively. The quantitative model also offers opportunity to generalize to finetune to control other metrics which don't necessarily have equivalent qualitative prompts like FRES.

\section{Conclusions}

This work introduces the readability-controlled text modification task.
Our task challenges controllable language models to generate eight versions of a text, each targeted for specific readability levels, in a manner independent from the source text readability.
Novel metrics, inspired by paraphrasing, assess the quality of readability-controlled text modification. Zero-shot adaptations for ChatGPT and Llama-2 show potential in steering readability but retain some correlation with the source text readability. A two-step process of generating paraphrases sequentially offers modest gains over one-step approaches.
Notably, more significant shifts in readability lead to reduced semantic and lexical similarity between source and target texts, highlighting the challenge of balancing readability control and content preservation.

\section{Ethics Statement}

There are no ethical concerns with this work.

\section{Limitations}

The main insights drawn from this work for controllable text modification are based upon a single dataset, CLEAR; some of the observations may not generalize to datasets in other domains. Additionally, the current work uses FRES as the readability function. There are alternative options too.

\section{Acknowledgements}

This research is funded by the EPSRC (The Engineering and Physical Sciences Research Council) Doctoral Training Partnership (DTP) PhD studentship and supported by Cambridge Assessment, University of Cambridge and ALTA.

\nocite{*}
\section{Bibliographical References}\label{sec:reference}

\bibliographystyle{lrec-coling2024-natbib}
\bibliography{lrec-coling2024-example}



\appendix

\section{Examples}

The below example uses ChatGPT 2-step to generate the paraphrases at each of the target readability scores.

\small

\begin{tcolorbox}[enhanced,attach boxed title to top center={yshift=-3mm,yshifttext=-1mm},
  colback=blue!5!white,colframe=blue!75!black,colbacktitle=red!80!black,
  title=Source,fonttitle=\bfseries,
  boxed title style={size=small,colframe=red!50!black} ]
When the young people returned to the ballroom, it presented a decidedly changed appearance. Instead of an interior scene, it was a winter landscape. The floor was covered with snow-white canvas, not laid on smoothly, but rumpled over bumps and hillocks, like a real snow field. The numerous palms and evergreens that had decorated the room, were powdered with flour and strewn with tufts of cotton, like snow. Also diamond dust had been lightly sprinkled on them, and glittering crystal icicles hung from the branches. At each end of the room, on the wall, hung a beautiful bear-skin rug. These rugs were for prizes, one for the girls and one for the boys. And this was the game. The girls were gathered at one end of the room and the boys at the other, and one end was called the North Pole, and the other the South Pole. Each player was given a small flag which they were to plant on reaching the Pole. This would have been an easy matter, but each traveller was obliged to wear snowshoes.
\newline\newline
\textbf{Readability:} 74.5
\end{tcolorbox}

\begin{tcolorbox}[enhanced,attach boxed title to top center={yshift=-3mm,yshifttext=-1mm},
  colback=blue!5!white,colframe=blue!75!black,colbacktitle=red!80!black,
  title=Target 5,fonttitle=\bfseries,
  boxed title style={size=small,colframe=red!50!black} ]
Upon the return of the youthful attendees to the grand hall, they noticed a conspicuous transformation had occurred. The traditional interior arrangement had been replaced with a wintry scene. The floor was covered with a pristine canvas resembling a snow-covered terrain, complete with irregular elevations and ripples. The palm trees and evergreen decorations in the room had also been transformed to resemble a snowy landscape, with fine grains of flour and fluffy clusters of cotton creating a snow-like effect. Additional decorations, such as diamond dust sprinkled on the flora, added a glittering icicle-like appearance. On opposite ends of the room, there were ethereal and beautifully displayed bear-skin rugs. These rugs were designated as coveted prizes for both the fair maidens and gallant gentlemen, creating a captivating rivalry. The game involved the young maidens gathering at one end of the room, while the young men gathered at the other. Referred to as the North Pole and South Pole, each participant was given a small flag as an accessory upon their arrival. Despite the seemingly simple task, achieving success in this quest proved to be a challenging endeavor. The participants had to navigate the terrain while wearing snowshoes, adding to the difficulty but also serving as a helpful aid. In summary, upon returning to the grand hall, the attendees were surprised to find a winter wonderland. The floor resembled a snowy landscape, and the decorations had been transformed to create a snowy effect. The bear-skin rugs at either end of the room served as prizes, and the participants had to navigate the terrain while wearing snowshoes.
\newline\newline
\textbf{Readability:} 41.9
\newline
\textbf{WER:} 127.6\%
\newline
\textbf{BERTScore F1:} 84.8\%
\newline
\textbf{Length change:} 47.5\%
\end{tcolorbox}

\begin{tcolorbox}[enhanced,attach boxed title to top center={yshift=-3mm,yshifttext=-1mm},
  colback=blue!5!white,colframe=blue!75!black,colbacktitle=red!80!black,
  title=Target 20,fonttitle=\bfseries,
  boxed title style={size=small,colframe=red!50!black} ]
Upon the return of the youthful attendees to the ballroom, a noticeable transformation had occurred. The environment had completely changed, resembling a wintry landscape. The dancing surface was covered in pristine, rumpled snow, imitating the uneven terrain of real snowfields. The palm trees and evergreen plants that had decorated the room were coated with flour and cotton, resembling snow. Diamond dust was delicately sprinkled to give an iridescent effect, while crystal icicles hung from the tree branches, adding to the overall charm.  On opposite walls of the room, there were two beautiful rugs made from the highly valued skins of bears. These rugs were meant as desirable prizes, with one given to the female participants and the other to the male contenders. This set the stage for the game, which engaged the participants in an exciting pursuit. The young ladies gathered at one end of the room, while the young gentlemen assembled at the other end, representing the North Pole and the South Pole, respectively. Each player was given a small flag, marking their goal of planting it at their designated pole.  However, this apparently simple task presented a significant challenge. Each adventurer had to wear snowshoes on their feet, adding a layer of complexity and skill to the endeavor.
\newline\newline
\textbf{Readability:} 47.7
\newline
\textbf{WER:} 96.0\%
\newline
\textbf{BERTScore F1:} 87.1\%
\newline
\textbf{Length change:} 17.3\%
\end{tcolorbox}

\begin{tcolorbox}[enhanced,attach boxed title to top center={yshift=-3mm,yshifttext=-1mm},
  colback=blue!5!white,colframe=blue!75!black,colbacktitle=red!80!black,
  title=Target 40,fonttitle=\bfseries,
  boxed title style={size=small,colframe=red!50!black} ]
Upon the return of the young people to the ballroom, they were met with a notable change. The previous interior had been transformed into a wintry landscape. The floor was covered in an uneven layer of white canvas, resembling real snow with its bumps and hills. The once-adorned palms and evergreens had been altered to look like snow-covered trees, with the addition of flour and cotton. These decorations were lightly sprinkled with diamond dust, and crystal icicles hung from the branches.  Magnificent bearskin rugs adorned the walls at each end of the room. These rugs were prizes to be awarded to one girl and one boy. The game involved the girls gathering at the North Pole and the boys assembling at the South Pole. Each participant was given a small flag to plant when they reached their destination.  However, this challenge wasn't as simple as it appeared, as every traveler had to wear snowshoes.
\newline\newline
\textbf{Readability:} 62.1
\newline
\textbf{WER:} 77.0\%
\newline
\textbf{BERTScore F1:} 88.1\%
\newline
\textbf{Length change:} -14.0\%
\end{tcolorbox}

\begin{tcolorbox}[enhanced,attach boxed title to top center={yshift=-3mm,yshifttext=-1mm},
  colback=blue!5!white,colframe=blue!75!black,colbacktitle=red!80!black,
  title=Target 55,fonttitle=\bfseries,
  boxed title style={size=small,colframe=red!50!black} ]
When the young people returned to the large room for the ball, there was a noticeable change. Instead of being inside, the room now looked like a wintry landscape. The floor was covered with a fabric that looked like snow, but it was not perfectly flat. Instead, it was crumpled and uneven, resembling a real snowfield. The various palm and evergreen trees that had originally decorated the room were coated with flour and had pieces of cotton attached to them to make them look like they were covered in snow. Also, a light sprinkle of diamond dust was put on, and shiny crystal icicles hung from the branches.  On the walls at each end of the room, there were two magnificent and beautiful rugs made from the skin of bears. These rugs were meant to be given as prizes - one for the girls and another for the boys. And this is how the game worked: the girls were positioned at one end of the room, while the boys were gathered at the opposite end. These ends were called the North Pole and the South Pole, respectively. Every player was given a small flag that they had to put in the ground when they reached their assigned end. However, there was a challenge involved - each participant had to wear special shoes that make it easier to walk on snow.
\newline\newline
\textbf{Readability:} 69.6
\newline
\textbf{WER:} 86.8\%
\newline
\textbf{BERTScore F1:} 88.8\%
\newline
\textbf{Length change:} 28.5\%
\end{tcolorbox}

\begin{tcolorbox}[enhanced,attach boxed title to top center={yshift=-3mm,yshifttext=-1mm},
  colback=blue!5!white,colframe=blue!75!black,colbacktitle=red!80!black,
  title=Target 65,fonttitle=\bfseries,
  boxed title style={size=small,colframe=red!50!black} ]
When the young people returned to the ballroom, it looked completely different. Instead of being indoors, it appeared to be a snowy winter scene. The floor was covered in bumpy and uneven white material, like a real snow field. The trees and plants in the room were made to look snowy with flour and bits of cotton, and had shiny ice hanging from them. There were nice bear-skin rugs on each wall of the room. These rugs were going to be given as prizes, one for the girls and one for the boys. Here's how the game worked: the girls stood on one side of the room and the boys stood on the other side. One side was called the North Pole and the other side was called the South Pole. Each person received a small flag that they had to stick into the floor when they reached their side. But here's the twist - everyone had to wear snowshoes!
\newline\newline
\textbf{Readability:} 81.6
\newline
\textbf{WER:} 64.9\%
\newline
\textbf{BERTScore F1:} 88.7\%
\newline
\textbf{Length change:} -10.6\%
\end{tcolorbox}

\begin{tcolorbox}[enhanced,attach boxed title to top center={yshift=-3mm,yshifttext=-1mm},
  colback=blue!5!white,colframe=blue!75!black,colbacktitle=red!80!black,
  title=Target 75,fonttitle=\bfseries,
  boxed title style={size=small,colframe=red!50!black} ]
When the young people returned to the ballroom, it looked completely different. Instead of being in a regular room, it looked like a scene from winter. The floor was covered in white cloth that resembled snow. It wasn't flat, but had bumps and hills like a real snow field. The plants and trees that had been used for decoration were covered in flour and bits of cotton to make them look like they were covered in snow. There was also some shiny material that looked like ice sprinkled on them, and there were shiny ice formations hanging from the branches.  On each side of the room, there were beautiful bear-skin rugs hanging on the walls. These rugs were going to be given as prizes, one for the girls and one for the boys. Now, let me explain the game they were going to play.  The girls were all standing on one side of the room, and the boys on the other. One side was called the North Pole, and the other was called the South Pole. Each player was given a small flag that they had to stick in the ground when they reached the Pole. But here's the catch - everyone had to wear snowshoes, which made it a bit more challenging.
\newline\newline
\textbf{Readability:} 80.9
\newline
\textbf{WER:} 75.3\%
\newline
\textbf{BERTScore F1:} 89.8\%
\newline
\textbf{Length change:} 19.0\%
\end{tcolorbox}

\begin{tcolorbox}[enhanced,attach boxed title to top center={yshift=-3mm,yshifttext=-1mm},
  colback=blue!5!white,colframe=blue!75!black,colbacktitle=red!80!black,
  title=Target 85,fonttitle=\bfseries,
  boxed title style={size=small,colframe=red!50!black} ]
When the young people returned to the ballroom, it looked very different. Instead of being inside, it seemed like a snowy winter scene. The floor was covered with bumpy white cloth that looked like snow, and the trees and plants were decorated with flour, bits of cotton, and shiny pieces that resembled ice. On the walls, there were two beautiful rugs made from bear skins. These rugs would be awarded as prizes, with one for the girls and one for the boys. Here's how the game worked: the girls were at one end of the room, called the North Pole, while the boys were at the other end, called the South Pole. Each player had to wear snowshoes and put a little flag in the ground when they reached their end of the room.
\newline\newline
\textbf{Readability:} 76.9
\newline
\textbf{WER:} 74.1\%
\newline
\textbf{BERTScore F1:} 87.4\%
\newline
\textbf{Length change:} -25.1\%
\end{tcolorbox}

\begin{tcolorbox}[enhanced,attach boxed title to top center={yshift=-3mm,yshifttext=-1mm},
  colback=blue!5!white,colframe=blue!75!black,colbacktitle=red!80!black,
  title=Target 95,fonttitle=\bfseries,
  boxed title style={size=small,colframe=red!50!black} ]
When the young people returned to the ballroom, it looked completely different. Instead of being inside, it looked like a snowy winter scene. The floor was covered in white fabric that looked like real snow. The trees and plants were covered in flour and bits of cotton to make them look snowy. They even had shiny ice hanging from the branches. On each wall, there was a beautiful bear-skin rug that would be given as a prize to the boys and girls. The game was for the girls to gather on one side and the boys on the other side. One side was called the North Pole and the other was the South Pole. Each person had to wear snowshoes and carry a small flag. The goal was to reach their assigned pole and plant their flag. This would have been easy, but the snowshoes made it a bit tricky.
\newline\newline
\textbf{Readability:} 82.5
\newline
\textbf{WER:} 75.3\%
\newline
\textbf{BERTScore F1:} 87.3\%
\newline
\textbf{Length change:} -16.2\%
\end{tcolorbox}

\end{document}